\documentclass[letterpaper,10pt]{article}
\usepackage[T1]{fontenc}
\usepackage{parskip}
\usepackage[font={small}]{caption}
\usepackage[hidelinks]{hyperref}
\usepackage[margin=2.5cm]{geometry}
\usepackage{times}
\usepackage{amsmath,amssymb,amsthm}
\usepackage[affil-it]{authblk}
\usepackage{listings}

\usepackage{microtype}
\usepackage{graphicx}
\usepackage{tikz}
\usepackage{subcaption}
\usepackage{booktabs} % for professional tables
\usepackage{amsmath}
\usepackage{amssymb}
\usepackage{verbatim}

\usepackage{natbib}

\usepackage{algorithm,algorithmic}

\usepackage{booktabs}         % Nice looking tables
\newcommand{\ra}[1]{\renewcommand{\arraystretch}{#1}}

\usepackage{hyperref}

\newcommand{\stepLength}{\alpha}

\renewcommand\mid{\,\vert\,}

\newcommand{\T}{n}
\newcommand{\R}{\mathbb{R}}

\newcommand{\N}{\mathcal{N}}
\newcommand{\msY}{\mathcal{Y}_1}
\newcommand{\meY}{\mathcal{Y}_2}
\newcommand{\msS}{\mathcal{S}_1}
\newcommand{\meS}{\mathcal{S}_2}

\newcommand{\Transp}{\mathsf{T}}
\newcommand{\tp}{\mathsf{T}}

\newcommand{\bmat}[1]{\begin{bmatrix} #1 \end{bmatrix}}

\newcommand{\myd}{\textrm{d}}      % Differential for integrals etc.

\date{}

\title{Stochastic quasi-Newton with adaptive step lengths for \newline large-scale problems}

\author[1]{Adrian Wills}
\author[2]{Thomas B. Sch\"on}
\affil[1]{School of Engineering, University of Newcastle, Australia. \newline
	Email: \texttt{adrian.wills@newcastle.edu.au}}
\affil[2]{Department of Information Technology, Uppsala University, Sweden. \newline
	Email: \texttt{thomas.schon@it.uu.se}}

\begin{document}

\maketitle

%%%%%%% Uncomment the following part for papers also published elsewhere
%\textbf{Please cite this version:}
%
%K. Karlsson, K. Kurtsson. A great title. In \textit{Proceedings of the 1st Semi-Annual Nordic Conference on Theory and Practice},
%Korpilombolo, Sweden, 2018.
%
%\begin{center}
%\begin{minipage}{.75\linewidth}
%\begin{lstlisting}[breaklines,basicstyle=\small\ttfamily]
%@inproceedings{TheGreatPaper,
%author    = {Karlsson, Kurt and Kurtsson, Karl},
%title     = {A great title},
%booktitle = {Proceedings of the 1st Semi-Annual Nordic Conference on Theory and Practice},
%year      = {2018},
%address   = {Korpilombolo, Sweden}
%}
%\end{lstlisting}
%\end{minipage}
%\end{center}
%
%\vspace{5em}
%%%%%%%%%%%%

\begin{abstract}
%	Sentence 1: A one-liner describing our key contribution.
	We provide a numerically robust and fast method capable of exploiting the local geometry 
%	curvature information 
	when solving large-scale stochastic optimisation problems.
%	Sentence 2-4: Elaborate on our construction.
	Our key innovation is an auxiliary variable construction coupled with an inverse Hessian approximation computed using a  receding history of iterates and gradients.
%	via a limited memory representation.
	It is the Markov chain nature of the classic stochastic gradient algorithm that enables this development. 
	The construction offers a mechanism for 
	stochastic line search adapting the step length.
%	choosing a potentially non-decreasing step length. 
% 	Sentence 5-6: Numerical results
	We numerically evaluate and compare against current state-of-the-art with encouraging performance on real-world benchmark problems where the number of observations and unknowns is in the order of millions.
\end{abstract}

%%%%%% Add also a pagebreak if you have uncommented the "Plase cite this version..."-stuff above
%\clearpage

%%%%%%%%%% The paper itself

%=============================================================================
%===============================  New Section  ===============================
%=============================================================================
\section{Introduction}
\label{sec:intro}
% Introduction
% 

%
%
%\begin{enumerate}
%	\item The first breakthrough in solving these stochastic optimization problems came almost 70 years ago. 
%	Key development, a Markov chain with decreasing increments.
%	\item Parallel development on deterministic problems, where state-of-the-art methods currently makes use of local second-order models and carefully computed increments using line-search algorithms.
%	\item There has been a significant amount of work of recent years on how to include second-order information (curvature) in computing the search direction, but significantly less when when it comes to adapting the step-length.
%	\item We provide a computationally cheap solution 
%	makes use of the curvature information
%	and 
%\end{enumerate}
%

% 0. A first paragraph stating what we will contribute and motivate the relevance of the problem
%
%\ts{Include a paragraph describing which problem we are interested in and how important it is. Briefly what we contribute.}

% 1. Proper statement of the problem we are interested in
% 
We are interested in the unconstrained \emph{stochastic} non-convex optimisation problem
\begin{subequations}
	\label{eq:problem}
	\begin{align}
		\min_{x\in\R^d}{f(x)}, %\quad \text{ where } f(x) = \frac{1}{n}\sum_{i=1}^{n}f_i(x),
	\end{align}
	when the cost function~$f(x)$ is on the form
	\begin{align}
		f(x) = \frac{1}{n}\sum_{i=1}^{n}f_i(x) + R(x),
	\end{align}
\end{subequations}
where~$d$ denotes the dimension of the unknown variable~$x$ and~$n$ denotes the number of available observations, i.e. the size of the dataset. Here, $f_i(x)$ denotes a loss function and $R(x)$ denotes a regularizer.
%We make use of~$d$ to denote the dimension of the unknown variable~$x$ and~$n$ denotes the number of available observations, i.e. the size of the dataset. We take a particular interest in situations where~$n$ and/or~$d$ are large.
The stochasticity of the problem is due to the fact that we only have access to \emph{noisy} evaluations of the cost function~$f(x)$ and its gradient~$\nabla f(x)$ according to
%\begin{subequations}
%	\label{eq:NoisyCostGrad}
	\begin{align}
		\label{eq:NoisyCostGrad}
		f_k = f(x_k) + e_k,\qquad %\\
%		\label{eq:GradMeas}
		g_k = \nabla f(x)|_{x = x_k} + v_k.
	\end{align}
%\end{subequations}
Here $e_k$ and $v_k$ denotes the noise on the function and gradient evaluations, respectively. 
We take a particular interest in situations where the number of data~$n$ and/or the number of unknowns~$d$ are vary large.

% 2. Explain why this is an important problem by pointing to different situations where its need arise
%
The stochastic optimisation problem~\eqref{eq:problem} is one of the most commonly encountered problems within supervised machine learning. The stochastic nature of the problem arises in different ways. 
%More concretely it arises for example in the following situations. 
First we mention large-scale problems where it is prohibitive to evaluate the cost function and its gradient on the entire dataset. Instead it is divided into several mini-batches via a subsampling procedure, which also explains where the noise arises. 
%		\ts{Marketing trick to get some results up front, which means that a self-explanatory caption important.} 
\begin{figure}[ht!]
	\centering
	\includegraphics[width=0.5\columnwidth]{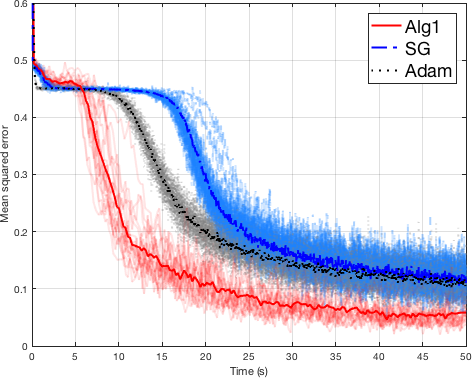}
	\caption{Solving the optimisation problem used in training a state-of-the-art deep convolutional neural network (CNN) used for recognizing images of handwritten digits from the MNIST data. Alg1 referes to our new developments in this paper, SG  refers to basic stochastic gradient and Adam refers to \cite{KingmaB:2015}. 
		For a full account of these experiments, see Section~\ref{sec:Exp}.}
	\label{fig:MNIST}
\end{figure}
As a second example we mention the use of numerical algorithms in approximately computing the cost function and its gradients, inevitably resulting in stochastic optimisation problems. We illustrate the result of our new developments on a problem of the first kind in Figure~\ref{fig:MNIST}, namely the optimisation problem arising in training a deep convolutional neural network. %solution to the MNIST problem of classifying images of hand-written images.
% Empirical risk minimization (ERM) problems occuring for example in deep learning, logistic regression, and support vector machines, etc.

% 3. Stress the Markov chain nature of SGD that is typically not used in recent developments even though it was clearly pointed out in the initial paper by Robbins and Monroe (1951).
% 
The first stochastic optimisation algorithm was introduced almost 70 years ago by \cite{RobbinsM:1951}. They made use of first-order information only, motivating the name stochastic gradient (SG) which is the contemporary machine learning term for these algorithms originally referred to as stochastic approximation. Interestingly most SG algorithms are not decent methods, since the stochastic nature of the update can easily produce a new iterate corresponding to an increase in the cost function, which is illustrated in Figure~\ref{fig:MNIST}. Instead they are in fact Markov chain methods, due to the fact that their update rule actually defines a particular Markov chain. This was indeed also clearly acknowledged already in the seminal paper by \cite{RobbinsM:1951}.
%, but after that it seems to have to a large extent been forgotten. 

% 4. State our key contributions
%
\textbf{Contributions and key properties:} We will heavily build upon the Markov chain nature of SG and our key contribution is a new construction enabled via an auxiliary variable trick allowing us to define an \emph{extended} Markov chain. The key feature of this construction is that we can efficiently make use of second-order (curvature) information in computing the search direction. This curvature information stems from an estimate of the inverse Hessian that we compute using a bounded history of previous iterates and stochastic gradients. The computational cost and memory footprint of this computation scales linearly in the number of data. 
Another important contribution is a stochastic line search capable of adapting the step length. From our numerical experiments we can see that this capability seems beneficial, especially in the beginning. 
% An interpretation is that our auxiliary variable construction effectively provides a stochastic line search. 
A practical feature is that our method only requires the user to select three tuning parameters, the size of the mini-batch, the size of the memory and the weight of a regulariser. 
We also develop a method for updating a Cholesky factor given the new measurement pair making our approach 
%The method is computationally cheap and numerically robust, which 
computationally cheap and numerically robust, which we illustrate using extensive numerical experiments comparing against current state-of-the-art methods on challenging large-scale real-world problems.

%=============================================================================
%===============================  New Section  ===============================
%=============================================================================
\section{Background and related work}
\label{sec:RW}
% Related work
% 
% The reason for naming this background and related work and not just related work is simply that 
% 

% 1. Basic ideas underlying both deterministic and stochastic methods 
%
Many numerical optimisation algorithms can be interpreted as learning algorithms, where the first step is to build a local model of the cost function~$f(x)$. This local model is then used to compute the next iterate, a new model is learned around this new iterate and the procedure is repeated. The so-called \emph{second-order} methods make use of quadratic Taylor series approximations $q_k(x)$ of $f(x)$ around the current iterate~$x_k$
\begin{align}
	\label{eq:2orderModel}
	q_k(x) &= f(x_k) + g_k^{\Transp}(x - x_k) 
	+ \frac{1}{2}(x - x_k)^{\Transp}H_k^{-1}(x - x_k),
\end{align}
where $g_k$ denotes an approximation of the gradient~$\nabla f(x_k)$ and $H_k$ denotes an approximation of the inverse Hessian $(\nabla^2f(x_k))^{-1}$. Direct minimisation of the quadratic model~\eqref{eq:2orderModel} suggests the following update of the iterates
\begin{align}
	\label{eq:GeneralUpdate}
	x_{k+1} = x_k - \stepLength_k H_k g_k,
\end{align}
where $\stepLength_k$ denotes the step length. The matrix~$H_k$ will be referred to as the \emph{scaling matrix} since it scales the gradient approximation~$g_k$. Many algorithms (including our present developments) update the iterates according to~\eqref{eq:GeneralUpdate}, but they differ greatly in how the components are found. Choosing the scaling matrix to be the identity $H_k = I$ we are back at the basic first-order gradient methods and with $H_k = (\nabla^2f(x_k))^{-1}$ we have Newton's method. The \emph{quasi-Newton} methods sit somewhere inbetween these two extremes, in that they employ a scaling matrix~$H_k$ that is a tractable approximation of the inverse Hessian. It is indeed this partial use of second-order information (curvature) that makes the quasi-Newton methods more robust and capable of reaching higher accuracy compared to pure gradient-based methods. The standard quasi-Newton method is the BFGS method, named after its inventors \citep{Broyden:1967,Fletcher:1970,Goldfarb:1970,Shanno:1970}. In its basic form this algorithm does not scale to the large-scale settings we are interested in. The idea of only making use of the most recent iterates and gradients in forming the inverse Hessian approximation was later suggested by \cite{Nocedal:1980} and \cite{LiuN:1989}. The result is a computationally cheaper method with a significantly reduced memory footprint, explaining the name L-BFGS, where the~L stands for limited memory. Due to its simplicity and good performance this has become one of the most commonly used second-order methods for large-scale problems. Our developments makes use of the same trick underlying L-BFGS, but it is carefully tailored to the stochastic setting. After this background let us now turn our attention to the most relevant related work when it comes to solving the stochastic problems we are interested in.

%======================================================
%======================================================
%===   This is where the related work really starts ===
%======================================================
%======================================================

% 1. Stochastic first order methods -- noise reduction methods 
%
%The standard first-order algorithms are described by for example \cite{} and \cite{BottouCN:2017}
The basic first-order SG algorithms have recently been significantly improved by the introduction of various noise reduction techniques, including the following methods; stochastic variance reduced gradient (SVRG) by \cite{JohnsonZ:2013}, Stochastic average gradient (SAG) \citep{SchmidtLB:2013}, Semi-Stochastic Gradient Descent (S2GD) \citep{KonecnyR:2017}, and SAGA \citep{DefazioBLJ:2014saga}. They all compute the gradient approximation via subsampling. There has recently also been some developments for non-convex settings, see e.g. \cite{ReddiHSPS:2016} and \cite{AllenZhyH:2016}. A thorough and forward-looking overview of the SG algorithm and its use within a modern machine learning context is provided by \cite{BottouCN:2017}. It also includes interesting accounts of possible improvements along the lines of first-order noise reduction techniques and second-order methods.

% 2. Stochastic second-order methods, in particular qN 
%
The well-known drawback of all first-order methods is that they do not make use of any curvature information. 
Analogously to the deterministic setting we can assemble methods that are numerically more robust and achieve better performance in general by also extracting and using second-order information, i.e. the curvature that is maintained in the form of the Hessian matrix or an approximation of it. Over the past decade we have witnessed  increasing capabilities of these so-called \emph{stochastic quasi-Newton methods}. There is still scope for significant developments when it comes to methods in this class and in this paper we aim to push the current boundaries.
%, which is also the category that our present developments belong to.

The work by \cite{SchraudolphYG:2007} developed modifications of BFGS and its limited memory version applicable to online stochastic optimisation problems. There has also been a series of papers approximating the scaling matrix~$H_k$ with a diagonal matrix, see e.g. \cite{BordesBG:2009} and \cite{DuchiEY:2011}. The idea of exploiting regularization together with BFGS was successfully introduced by \cite{MokhtariR:2014}, where the scaling matrix~$H_k$ was  modified using regularization. Later they \citep{MokhtariR:2015} also developed a stochastic L-BFGS algorithm without regularization. The idea of replacing the stochastic gradient difference in the BFGS update with a subsampled Hessian-vector product was recently introduced by \cite{ByrdHNS:2016} and \cite{WangMGL:2017} introduced a damped L-BFGS method.

%
%\ts{How are we different?: Important difference to our work, we use variable step size analogously to what is done for deterministic problems, whereas all previous work---to the best of the authors knowledge---are forced to using a step size~$\alpha_k$ that decrease with~$k$.}
%

% 3. Recent combinations of stochastic second-order methods and first-order noise reduction methods
% 
Over the past five years we have also seen quite a lot of fruitful activity in combining the stochastic quasi-Newton algorithms with various first-order noise reduction methods. \cite{MoritzNJ:2016} successfully showed that it is possible to combine the L-BFGS methods by \cite{ByrdHNS:2016} with the SVRG noise reduction algorithm by \cite{JohnsonZ:2013} to reduce the problem with noisy gradients. Along this line of work we also find \cite{GowerGR:2016} where the authors introduced a stochastic block BFGS update that they then combined with the SVRG method.

%
%\ts{exactly what is meant by a block update and how does it relate to our development?}
%

%
Contrary to almost all of the existing work mentioned above we make explicit use of and build upon the fact that the SG algorithm is a particular Markov chain designed specifically to solve the stochastic optimisation problem.

% 4. MCMC type approaches - the second strand of related work
%
Related to the Markov chain theme, the highly innovative work by \cite{WellingT:2011} has recently sparked a relevant parallel development within the Markov chain Monte Carlo (MCMC) literature for the case when $f(x)$ can be interpreted as a likelihood function. The aim is to exploit the geometry of the target  distribution (the posterior) by using constructions from stochastic optimisation and Langevin diffusion dynamics. The use of a carefully designed local curvature estimate was enabled by \cite{SimsekliBCR:2016} when they incorporated ideas from L-BGFS within an MCMC setting. The main focus of this MCMC work has been directed towards exploring the posterior distribution when the chain is initialised at a ``good'' initial point (e.g. \cite{teh2016consistency} assume a MAP estimate to start the chain). In contrast, here we are primarily interested in rapid convergence towards an area of minimum cost from any initial point and for a more general class of cost functions.

%=============================================================================
%===============================  New Section  ===============================
%=============================================================================
\section{Algorithm aummary} % algorithm}
\label{sec:Alg}
% New Algorithm
%
%

%
%\ts{An idea for the structure of this section:
%\begin{enumerate}
%	\setlength{\itemsep}{-1mm}	
%	\item Resulting algorithm
%	\item Computing the search direction
%	\item "Auxiliary variable construction enabling stochastic line search"
%%	\item Reducing the bias via independent sampling (propably better to leave this for the analysis, especially since it does not seem to be as important as we first thought.)
%\end{enumerate}
%}

The key innovation in our solution lies in an auxiliary variable
construction allowing for line search within a stochastic quasi-Newton
setting. Hence, we are no longer forced to make use of decreasing step
lengths in solving stochastic optimisation problems. As can be seen in
Algorithm~\ref{alg:SQN} the overall structure of our solution is
similar to most existing solutions, but all details have been
carefully tailored to the stochastic setting. We start by describing
how the search direction is calculated (rows 4-5) in
Section~\ref{sec:SearchDir}. Here, we take care to derive a
numerically robust and fast update of the inverse Hessian
approximation. The auxiliary variables construction (rows 7-9)
described in Section~\ref{sec:Analysis} allows for the use of step
lengths that adapt according the local geometry, resulting in a
functionality very similar to standard deterministic second-order
algorithms with line search.

%
%Our algorithm follows the overall structure of many of the most existing stochastic algorithms, but there is a key difference is that we have introduced an accept/reject stage (rows 7-9 in Algorithm~\ref{alg:SQN}). 

\begin{algorithm}[htb!]
  \caption{\textsf{Stochastic quasi-Newton with line search}}
  \begin{algorithmic}[1]
    \REQUIRE An initial estimate $x_1$, a maximum number of iterations
    $k_{\max}$ and maximum step-length
    $0 < \bar{\alpha}_{k} \leq 1$. Choose $\rho \in \{ 0,1\}$, where
    $\rho = 1$ provides SG decay rate on step length $\alpha_k$, and $\rho = 0$ guarantees that the
    step-length will not exceed $\bar{\alpha}_{k}$. Choose a step-length
    scaling factor $\kappa \in (0,1)$.%
    \STATE Set $k = 1$ and $\alpha_1 = \bar{\alpha}_{1}$ and perform the following.
    \WHILE{$k < k_{\max}$}
%    \vspace{5mm}
    \STATE \textbf{Search direction calculation:}
    \STATE \label{step1} Obtain a measurement of the cost function and its gradient
	\vspace{-5mm}
    \begin{subequations}
      \label{eq:1}
      \begin{align}
        f_k &= f(x_k) + e_k,\\
        g_k &= \nabla f(x_k) + v_k.
      \end{align}
    \end{subequations}
    \STATE Calculate a search direction $p_k$ such that 
    \begin{align}
      \label{eq:2}
      \begin{cases}
        p_k^{\Transp}g_k < 0, & \|g_k\| > 0,\\
        p_k = 0, & \text{otherwise}.
      \end{cases}
    \end{align}
%    \vspace{5mm}
    \STATE \textbf{New iterate calculation:}
    \STATE Compute proposal $\xi_{k+1} = x_k+\alpha_k p_k$.
    \STATE Calculate the acceptance indicator variable
    \begin{align}
      \label{eq:12}
      c_k &= \begin{cases} 
        1, & \text{w.p.}\quad \max \{\rho\ ,\ a(\xi_{k+1} \mid
        x_k) \},\\
        0, & \text{otherwise}.
        \end{cases}
    \end{align}
    \STATE Update the variables
	\begin{subequations}
          \label{eq:13}
          \begin{align}
            x_{k+1}      &= x_k + c_k \alpha_k p_k,\\
            p_{k+1}      &= p_k,\\
            \alpha_{k+1} &= c_k \left (\frac{1}{k} \right)^\rho \bar{\alpha}_{k} + (1-c_k)\kappa \alpha_k.
          \end{align}
        \end{subequations}
        \IF{$c_k = 0$} 
        \STATE Set $k \leftarrow k+1$ and return to step~7.
        \ELSE
        \STATE Set $k \leftarrow k+1$ and return to step~2.
        \ENDIF
    % \STATE Set $\alpha = \alpha_{\max}$ and set an indicator variable
    % $c = 0$ and perform the following.
    % \WHILE{$\alpha \geq \alpha_{\min}$ \AND $c = 0$}
    % \STATE Compute the candidate sample 
    % \begin{align}
    %   \label{eq:3}
    %   \xi_k = x_k + \alpha p_k.
    % \end{align}
    % \STATE Compute the acceptance probability 
    % \begin{align}
    %   \label{eq:4}
    %   a(\xi_k \mid x_k) \in (0,1].
    % \end{align}
    % \STATE Draw a random number $u_k \sim \mathcal{U}(0,1) \in [0,1]$.
    % \IF{$u_k < a(\xi_k \mid x_k)$}
    % \STATE Set $x_{k+1} = \xi_k$.
    % \STATE Set $k \leftarrow k+1$.
    % \STATE Set $c = 1$. 
    % \ELSE
    % \STATE Set $x_{k+1} = x_k$.
    % \STATE Update $\alpha \leftarrow \kappa \alpha$.
    % \STATE Set $k \leftarrow k+1$.
    % \ENDIF
    % \ENDWHILE
    \ENDWHILE
  \end{algorithmic}
  \label{alg:SQN}
\end{algorithm}

%=============================================================================
%===============================  New Section  ===============================
%=============================================================================
\section{Search direction computation} % algorithm}
\label{sec:SearchDir}
%=============================================================================
%===============================  New Section  ===============================
%=============================================================================
%\subsection{Computing the Search Direction $p_k$}

In this section we address the problem of computing a search direction based on having a limited memory available for storing previous gradients and associated iterates. The approach we adopt is similar to limited memory quasi-Newton methods, but here we employ a direct least-squares estimate of the inverse Hessian matrix rather than more well-known methods such as damped L-BFGS and L-SR1. The main reason for considering the least-squares approach is that it appears to perform quite well against the alternative methods for the class of problems considered in this paper. We construct a limited-memory inverse Hessian approximation in Section \ref{sec:inverseHess} and show how to update this representation in Section~\ref{sec:fastNrobust}. Section~\ref{sec:descent} provides a means to ensure that a descent direction is calculated.

\subsection{Inverse Hessian approximation}
\label{sec:inverseHess}
According to the Secant condition (see e.g. \cite{Fletcher:1987}), the inverse Hessian matrix $H_k$ should satisfy 
\begin{align}
  H_k y_k = s_k,
\end{align}
where $y_k = g_k - g_{k-1}$ and $s_k = x_k - x_{k-1}$. Since there are  generally more unknown values in $H_k$ than can be determined from $y_k$ and $s_k$ alone, quasi-Newton methods update $H_k$ from a previous estimate by solving problems of the type
%\begin{align}
%  H_k = \arg \min_{H} \|H - H_{k-1}\|^2_F, \quad \text{s.t.} \quad H=H^{\Transp}, \quad Hy_k = s_k.
%\end{align}
\begin{equation}
	\begin{aligned}
	H_k = \arg \min_{H} \quad &  \| H - H_{k-1} \|^2_{F,W}\\
	\text{s.t.} \quad &  H=H^{\Transp}, \quad Hy_k = s_k,
	\end{aligned}
\end{equation}
where $\|X\|^2_{F,W} = \|XW\|^2_F = \text{trace}(W^\tp X^\tp X W)$ and the
choice of weighting matrix $W$ results in different algorithms (see
\cite{Hennig:2015} for an interesting perspective on this).

Here we employ a similar approach and determine $H_k$ as the solution
to the following regularised least-squares problem
\begin{align}
  \label{eqn:Hupdate}
  H_k = \arg \min_{H} \|HY_k - S_k\|^2_F + \lambda \|H - \bar{H}_k\|^2_F,
\end{align}
where $Y_k$ and $S_k$ hold a limited number of past $y_k$'s and $s_k$'s according to
\begin{subequations}
	\begin{align}
		Y_k &\triangleq \bmat{y_{k-m+1},\ldots,y_k}, \\
		S_k &\triangleq \bmat{s_{k-m+1}, \ldots,s_k},
	\end{align}
\end{subequations}
and $m << n$ is the memory limit. The regulator matrix $\bar{H}_k$
acts as a prior on $H$ and can be modified at each iteration $k$. The
parameter $\lambda >0 $ is used to control the relative cost of the
two terms in \eqref{eqn:Hupdate}. It can be verified that the
solution to the above least-squares problem (\ref{eqn:Hupdate}) is
given by
\begin{align}
	\label{eq:Hupdate}
	H_k = \left ( \lambda I + Y_kY_k^{\Transp} \right )^{-1} \left
  ( \lambda  \bar{H}_k + Y_k S_k^{\Transp}\right ),
\end{align}
where $I$ denotes the identity matrix. The above inverse Hessian estimate can be used to generate a search direction in the standard manner by scaling the negative gradient, that is
\begin{align}
\label{eq:newAlgorithm:2}
  p_k = -H_k g_k.
\end{align}
However, for large-scale problems this is not practical since it
involves the inverse of a large matrix. To ameliorate this difficulty,
we adopt the standard approach by storing only a minimal (limited
memory) representation of the inverse Hessian estimate $H_k$. To
describe this, note that the dimensions of the matrices involved are
\begin{align}
  H_k \in \R^{d \times d}, \qquad
  Y_k \in \R^{d \times m}, \qquad
  S_k \in \R^{d \times m}.
\end{align}
We can employ the Sherman--Morrison--Woodbury formula to arrive at the following equivalent expression for~$H_k$
\begin{align}
  H_k &= \left [ I - Y_k \left ( \lambda I + Y_k^{\Transp}Y_k \right )^{-1}
        Y_k^{\Transp} \right ] \left ( \bar{H}_k + \lambda^{-1} Y_k S_k^{\Transp}\right ).
\end{align}
Importantly, the matrix inverse
$\left ( \lambda I + Y_k^{\Transp}Y_k \right )^{-1}$ is now by
construction a positive definite matrix of size $m \times
m$. Therefore, we will construct and maintain a Cholesky factor of
$I + Y_k^{\Transp}Y_k$ since this leads to efficient solutions. In
particular, if we express this matrix via a Cholesky decomposition
\begin{align}
  R_k^{\Transp} R_k &= \lambda I + Y_k^{\Transp}Y_k,
\end{align}
where $R_k \in \R^{m \times m}$ is an upper triangular matrix, then the search direction $p_k = -H_kg_k$ can be computed via
\begin{subequations}
	\begin{align}
		p_k &= -z_k  + Y_k w_k,\\
		z_k &= \bar{H}_k g_k + \lambda^{-1} Y_k(S_k^{\Transp}g_k),\\
		w_k &= R_k^{-1} \left ( R_k^{-\Transp} \left ( Y_k^{\Transp} z_k\right ) \right ).
	\end{align}
\end{subequations}
% \ts{Notation detail: $v_k$ used for gradient noise. Not a big deal.}
Constructing $R_k$ can be achieved in several ways. The so-called
normal-equation method constructs the (upper triangular) part of
$\lambda I + Y_k^{\Transp}Y_k$ and then employs a Cholesky routine, which produces $R_k$ in $O(n\frac{m(m+1)}{2} + m^3/3)$ operations. Alternatively, we can compute $R_k$ by applying Givens rotations or Householder reflections to the matrix 
\begin{align}
	M_k = \bmat{\sqrt{\lambda} I \\ Y_k}.
\end{align}
This costs $O(2m^2((n+m) - m/3)$ operations, and is therefore more expensive, but typically offers better numerical accuracy \citep{GolubVL:2012}.

%=============================================================================
%===============================  New Section  ===============================
%=============================================================================
\subsection{Fast and robust inclusion of new measurements}
\label{sec:fastNrobust}
In order to maximise the speed, we have developed a method for updating a Cholesky factor given the new measurement pair $(s_{k+1}, y_{k+1})$. Suppose we start with a Cholesky factor~$R_k$ at iteration~$k$ such that 
\begin{align}
  \label{eq:5}
  R_k^{\Transp} R_k &= \lambda I + Y_k^{\Transp}Y_k
\end{align}
and that we are given a new measurement pair $(s_{k+1},y_{k+1})$. Assume, without loss of generality, that $Y_k$ and $S_k$ are ordered in the following manner
\begin{subequations}
	\begin{align}
		\label{eq:6}
		Y_k &\triangleq \bmat{\msY, y_{k-m+1}, \meY},\\
		S_k &\triangleq \bmat{\msS, s_{k-m+1}, \meS},
\end{align}
\end{subequations}
where $\msY$, $\meY$, $\msS$ and $\meS$ are defined as
\begin{subequations}
	\begin{align}
	  \label{eq:8}
	  \msY &\triangleq \bmat{y_{k-m+\ell+1}, \ldots, y_k},\\
	  \meY &\triangleq \bmat{y_{k-m+2},\ldots,y_{k-m+\ell}},\\
	  \msS &\triangleq \bmat{s_{k-m+\ell+1}, \ldots, s_k},\\
	  \meS &\triangleq \bmat{s_{k-m+2},\ldots,s_{k-m+\ell}},
\end{align}
\end{subequations}
and $\ell$ is an appropriate integer so that $Y_k$ and $S_k$ have $m$ columns. The above ordering arises from ``wrapping-around'' the index when storing the measurements. We create the new $Y_{k+1}$ and $S_{k+1}$ by replacing the oldest column entries, $y_{k-m+1}$ and $s_{k-m+1}$, with the latest measurements $y_{k+1}$ and $s_{k+1}$, respectively, so that 
\begin{subequations}
\begin{align}
  \label{eq:7}
  Y_{k+1} &\triangleq \bmat{\msY, y_{k+1}, \meY},\\
  S_{k+1} &\triangleq \bmat{\msS, s_{k+1}, \meS},
\end{align}
\end{subequations}
The aim is to generate a new Cholesky factor $R_{k+1}$ such that 
\begin{align}
  \label{eq:9}
  R_{k+1}^{\tp} R_{k+1} &= \lambda I + Y_{k+1}^{\tp}Y_{k+1}.
\end{align}
To this end, let the upper triangular matrix $R_k$ be written
conformally with the columns of $Y_k$ as
\begin{align}
  \label{eq:10}
  R_k = \bmat{\mathcal{R}_1 & r_1 & \mathcal{R}_2 \\ & r_2 & r_3 \\ &
                                  & \mathcal{R}_4}
\end{align}
so that $\mathcal{R}_1$ and $\mathcal{R}_2$ have the same number of
columns as $\msY$ and $\meY$, respectively. Furthermore, $r_1$ is a column
vector, $r_2$ is a scalar and $r_3$ is a row vector. Therefore,
\begin{align}
  \label{eq:11}
  R_k^\tp &R_k = \bmat{\mathcal{R}_1^\tp \mathcal{R}_1 
  & \mathcal{R}_1^\tp r_1 &
  \mathcal{R}_1^\tp \mathcal{R}_2 \\
  \cdot & r_2^2 + r_1^\tp r_1 & r_1^\tp \mathcal{R}_2 + r_2r_3 \\ 
          \cdot & \cdot & \mathcal{R}_4^\tp \mathcal{R}_4 +
                          \mathcal{R}_2^\tp \mathcal{R}_2 + r_3^\tp
                          r_3}\nonumber\\
  &= \bmat{\lambda I+\msY^\tp\msY & \msY^\tp y_{k-m+1} & \msY^\tp \meY\\
  \cdot & \lambda + y_{k-m+1}^\tp y_{k-m+1} & y_{k-m+1}^\tp \meY \\
  \cdot & \cdot & \lambda I + \meY^\tp \meY}
\end{align}
By observing a common structure for the update $\lambda I + Y_{k+1}^\tp
Y_{k+1}$ it is possible to write
\begin{align}
   \label{eq:14}
  &\lambda I + Y_{k+1}^\tp Y_{k+1} \nonumber\\
  &=\bmat{\lambda I+\msY^\tp\msY & \msY^\tp y_{k+1} & \msY^\tp \meY\\
  \cdot & \lambda + y_{k+1}^\tp y_{k-m+1} & y_{k+1}^\tp \meY \\
  \cdot & \cdot & \lambda I + \meY^\tp \meY}\nonumber\\
  &= \bmat{\mathcal{R}_1^\tp \mathcal{R}_1 
  & \mathcal{R}_1^\tp r_4 &
  \mathcal{R}_1^\tp \mathcal{R}_2 \\
  \cdot & r_5^2 + r_4^\tp r_4 & r_4^\tp \mathcal{R}_2 + r_5r_6 \\ 
          \cdot & \cdot & \mathcal{R}_6^\tp \mathcal{R}_6 +
                          \mathcal{R}_2^\tp \mathcal{R}_2 + r_6^\tp
                          r_6}
\end{align}
where $r_4$, $r_5$ and $r_6$ are determined by
\begin{subequations}
\begin{align}
  \label{eq:15}
  r_4 &= \mathcal{R}_1^{-\tp} (\msY^\tp y_{k+1}),\\
  r_5 &= \left ( \lambda + y_{k+1}^\tp y_{k+1} - r_4^\tp r_4 \right
        )^{1/2},\\
  r_6 &= \frac{1}{r_5} \left ( y_{k+1}^\tp \meY - r_4^\tp
        \mathcal{R}_2 \right ).
\end{align}
\end{subequations}
The final term $\mathcal{R}_6$ can be obtained by noticing that 
\begin{align}
  \label{eq:16}
  \mathcal{R}_6^\tp \mathcal{R}_6 + \mathcal{R}_2^\tp \mathcal{R}_2 + r_6^\tp r_6 
  &= \mathcal{R}_4^\tp \mathcal{R}_4 + \mathcal{R}_2^\tp \mathcal{R}_2 + r_3^\tp r_3,
\end{align}
implies
\begin{align}
  \label{eq:17}
  \mathcal{R}_6^\tp \mathcal{R}_6 & = \mathcal{R}_4^\tp \mathcal{R}_4 - r_6^\tp r_6 + r_3^\tp r_3.
\end{align}
Therefore $\mathcal{R}_6$ can be obtained in a computationally very efficient manner by down-dating and updating the Cholesky factor
$\mathcal{R}_4$ with the rank-1 matrices $r_6^\tp r_6$ and
$r_3^\tp r_3$, respectively (see e.g. Section 12.5.3 in  \cite{GolubVL:2012}).
% \begin{algorithm}[htb!]
%   \caption{\textsf{Cholesky Factor Update}}
%   \begin{algorithmic}[1]
%     \REQUIRE The memory size $m$, the current iteration number $k$ and the latest measurements $y_k$ and $s_k$.
%     \IF{$k=1$}
%     \STATE Do something.
%     \ELSIF{$k<m$}
%     \STATE Do something.
%     \ELSE
%     \STATE Do something.
%     \ENDIF
%   \end{algorithmic}
%   \label{alg:UpdateCholesky}
% \end{algorithm}

%=============================================================================
%===============================  New Section  ===============================
%=============================================================================
\subsection{Ensuring a descent direction}% and Choosing $\bar{H}_k$}
\label{sec:descent}
In Algorithm~\ref{alg:SQN} we stipulate that the search direction
$p_k$ must be chosen to mimic a descent direction such that
$p_k^\tp g_k < 0$. Due to the fact that the gradient is not exact,
then this descent condition does not strictly enforce a descent direction, but
it is nonetheless useful to satisfy the descent condition in practice. The
search direction~$p_k$ as determined by~\eqref{eq:newAlgorithm:2} will
not be a descent direction in general since the approximation $H_k$ of
the inverse Hessian is not necessarily positive
definite. Nevertheless, by observing that
\begin{align}
  \label{eq:newAlgorithm:1}
  g_k^\tp (p_k + \beta g_k) = g_k^\tp p_k - \beta g_k^\tp g_k,
\end{align}
we can always choose a $\beta \geq 0$ such that $p_k + \beta g_k$ is a descent 
direction with respect to the inexact gradient~$g_k$. For example, we can choose 
\begin{align}
  \label{eq:newAlgorithm:3}
  \beta = 2 \max \left \{ 0, \frac{p_k^\tp g_k}{g_k^\tp g_k} \right \}.
\end{align}
It is also worth pointing out that this situation occurred very
infrequently during all of the experiments reported in
Section~\ref{sec:Exp}. The above is by no means an optimal strategy,
but it appears to perform very well in practice.

% In terms of choosing the regularising matrix $\bar{H}_k$ it proved
% very useful to employ a simple strategy of
% \begin{align}
%   \label{eq:newAlgorithm:4}
%   \bar{H}_k \triangleq \gamma_k I
% \end{align}
% where the positive scalar $\gamma_k >0$ was adaptively chosen
% according to recent progress. As a crude measure of progress we
% adopted the following rule
% \begin{align}
%   \label{eq:newAlgorithm:5}
%   \gamma_k = \begin{cases}
%     \kappa \gamma_k
%   \end{cases}
% \end{align}
% \aw{We change the above search direction to ensure that the inner
%   product is negative}.
% \aw{We also adapt the regularizer matrix $\bar{H}_k$ according to
%   recent progress}

%=============================================================================
%===============================  New Section  ===============================
%=============================================================================
\section{Auxiliary variable construction}
\label{sec:Analysis}
% Analysis
%
%

% 
%\subsection{Auxiliary variable construction enabling stochastic line search}
%\label{sec:AVC}
Algorithm~\ref{alg:SQN} offers two distinct variants. If the parameter
$\rho=1$, then the algorithm will mimic a classical SG approach in
that we accept every proposal~$\xi_{k+1}$ according to~\eqref{eq:12} 
and we are free to choose $\bar{\alpha}_{k}$ as a decaying sequence
\begin{align}
  \label{eq:newAlgorithm:6}
  \bar{\alpha}_{k} \triangleq \frac{\bar{\alpha}_0}{k}, \quad \text{for
  some fixed } \bar{\alpha}_0 > 0.
\end{align}
Therefore, $\alpha_{k+1} = \bar{\alpha}_0/k$, which is a typical
choice for many SG algorithms. In this case we can employ all the
analysis from SG methods, see \cite{BottouCN:2017}. 

The alternative $\rho = 0$, offers a different approach, which is our
main focus in this work as detailed in Section~\ref{sec:ASL}. Our algorithm
produces a Markov chain and in Section~\ref{sec:EPE} it is briefly
described how we can use it to extract a competitive point estimate.

%=============================================================================
%===============================  New Section  ===============================
%=============================================================================
\subsection{Adaptive step lengths}
\label{sec:ASL}
%
% Here the acceptance probability is calculated as 
% \begin{subequations}
%   \label{eq:newAlgorithm:7}
% 	\begin{align}
%   a(\xi_{k+1},x_k) &= \begin{cases} 
%     1 & \epsilon_k > 0,\\
%     \mathcal{C}(-\epsilon_k,\sigma^2) & \text{otherwise}.
%   \end{cases}\\
%   \epsilon_k &\triangleq f(\xi_{k+1})-f(x_{k}),
% \end{align}
% \end{subequations}
% where $\mathcal{C}(-\epsilon_k,\sigma^2)$ denotes the cumulative distribution 
% function for a Gaussian with mean~$-\epsilon_k$ and variance~$\sigma^2$. 
%
%
%
%Therefore, we discuss the case where $\rho=0$ in the following.
%=============================================================================
%===============================  New Section  ===============================
%=============================================================================
%\subsection{Convergence in distribution}
When we set~$\rho = 0$ in Algorithm~\ref{alg:SQN} it will generate an $m^{\text{th}}$-order Markov chain
$\{x_{k-m+1:k}, \alpha_{k-m+1:k}, u_{k-m+1:k}\}_{k\geq 1}$ where the
notation $x_{k-m+1:k} \triangleq \{x_{k-m+1}, \ldots, x_k\}$ is used
to represent the past $m$ iterates. The first auxiliary variable~$\alpha_k$
is the step length from Algorithm~\ref{alg:SQN} and the second
auxiliary variable $u_k$ represents the information
required to evaluate the approximate (noisy) cost and gradient. For
example, in the case of subsampling, $u_k$ represents the subset of
integers from $\{1, \ldots,n\}$ used to approximate the subsampled
cost and associated gradient. In Sequential Monte Carlo (SMC)  methods (used in Section~\ref{sec:SMC}), the auxiliary variable $u_k$ represents the selection of modes that propagate through the filter in order to again estimate the likelihood and its gradient (see \cite{AndrieuDH:2010} for details).

In what follows, we make the dependence on the auxiliary variable $u_k$
explicit by using the notation that $f(x_k,u_k)$ is the cost
approximation and $g(x_k,u_k)$ is the gradient of $f(x_k,u_k)$ with
respect to $x$.

The Markov chain evolves according to 
\begin{subequations}
  \label{eq:17}
  \begin{align}
    \label{eq:11}
    x_{k+1} &= x_k + c_k \alpha_k p_k,\\
    p_k    &= -H_kg_k - 2 \max  \left \{ 0,\frac{g_k^\tp H_k g_k}{g_k^\tp
             g_k} \right \} g_k,\\
    g_k    &= g(x_k,u_k),\\
    H_k    &= H(x_{k-m+1:k},\alpha_{k-m+1:k}, u_{k-m+1:k}),\\
    \alpha_{k+1} &= c_k + (1-c_k)\kappa \alpha_k, \label{eq:16}
  \end{align}
\end{subequations}
where $H(x_{k-m+1:k},\alpha_{k-m+1:k}, u_{k-m+1:k})$ is defined as
$H_k$ in \eqref{eq:Hupdate}, but here we highlight that the inverse
Hessian approximation is a function of the past $m$ iterates
$x_{k-m+1:k}$ and of the auxiliary variables $u_{k-m+1:k}$ and
$\alpha_{k-m+1:k}$ over this same window. The variable $c_k$ is
determined by
\begin{align}
	\label{eq:14}
	c_k &= \begin{cases} 
    1, & \text{w.p.}\quad a(x_k+\alpha_k p_k \mid x_k),\\
    0, & \text{otherwise},
\end{cases}
\end{align}
where the acceptance probability is calculated as 
\begin{subequations}
  \label{eq:newAlgorithm:7}
	\begin{align}
  a(\xi_{k+1},x_k) &= \begin{cases} 
    1 & \epsilon_k < 0,\\
    \mathcal{C}(-\epsilon_k,\sigma^2) & \text{otherwise}.
  \end{cases}\\
  \epsilon_k &\triangleq f(\xi_{k+1})-f(x_{k}),
\end{align}
\end{subequations}
where $\mathcal{C}(-\epsilon_k,\sigma^2)$ denotes the cumulative
distribution function for a Gaussian with mean~$-\epsilon_k$ and
variance~$\sigma^2$. The acceptance probability
in~\eqref{eq:newAlgorithm:7} has the effect of strictly accepting
proposals that decrease the cost, while accepting those that increase
the cost with a probability
$\mathcal{C}(-\epsilon_k,\sigma^2)$. Therefore, a proposed $\xi_{k+1}$
that causes a large increase in the cost, relative to the uncertainty
of the cost, is very unlikely to be accepted. Note that it is possible
to readily calculate an unbiased estimate of the cost function
variance~$\sigma^2$, and this can be re-evaluated as the algorithm
progresses.

Should a proposal be rejected then the step length is reduced
according to $\alpha_{k+1} = \kappa \alpha_k$ and the algorithm
returns to proposing a new~$\xi_{k+1}$ with reduced step length in
Step~7 without calculating a new search direction (the intent is
similar to stochastic line search algorithms
\citep{MahsereciH:2017}). In the event that the proposal is accepted
then $\alpha_{k+1} = \bar{\alpha}_{k}$, which for this variant of the
algorithm was chosen as $\bar{\alpha}_{k} = 1$ for all~$k$.

\textbf{Comments:} 
A natural question to ask is that of convergence of the proposed
algorithm. Convergence of a Markov chain to an invariant distribution has been the subject of intense research within  statistics and
related communities, see e.g. \cite{MeynT:2009} for a solid textbook account. Essentially, if it can be shown that the Markov transition kernel is invariant, that the chain is irreducible, and that it is also aperiodic, then it will converge to a stationary distribution. However, it is not immediately obvious (or indeed possibly correct) to assert that the transition kernel devised in Algorithm~\ref{alg:SQN} is invariant. 
\subsection{Extracting estimates}
\label{sec:EPE}
As discussed above, Algorithm~\ref{alg:SQN} produces
iterates~$\{x_k\}_{k\geq 1}$ that are distributed according to some
underlying distribution $p(x)$, that in accordance with the acceptance
probability, favours reductions in the cost function. As with standard
Markov chain methods, we can then utilise these samples via a law of
large numbers argument to form expectations of the type
\begin{align}
  \label{eq:18}
  h = \int h(x) p(x) dx = \lim_{M \to \infty} \frac{1}{M} \sum_{k=1}^M h(x_k),
\end{align}
where $h(\cdot)$ refers to a test function. The utility of this approach is that we can produce as many samples
from the target distribution as required in order to compute a desired
expectation.

In the experiments presented in Section~\ref{sec:Exp}, we
employed a very simple strategy of computing the expected value of
$x$, so that $h(x) = x$, which results in the following estimate
\begin{align}
  \label{eq:newAlgorithm:8}
  \widehat{x} = \frac{1}{M} \sum_{k=k_{\min}}^{M+k_{\min}-1} x_k
\end{align}
where $k_{\min} > 0$ defines a minimum number of transient iterations to ignore
in the calculation. The results summarised in Table~\ref{tab:table2}
were calculated according to \eqref{eq:newAlgorithm:8} by using the
final 20\% of the iterations.

\section{Numerical experiments}
\label{sec:Exp}
% Experiments
% 
%

Let us now put our new developments to test on a suite of problems from four different categories carefully chosen to exhibit different properties and challenges. In Section~\ref{sec:SE} we study a synthetic example to gauge the performance in a controlled setting. 
We then move on to more interesting and challenging problems involving large-scale and real-world data. In particular we will in Section~\ref{sec:MNIST} consider an optimisation problem arising from the use of deep learning to solve the classical machine learning benchmark MNIST\footnote{\url{yann.lecun.com/exdb/mnist/}}, where the task is to classify images of handwritten digits. 
Another commonly used benchmark is considered in Section~\ref{sec:LIBSVM}, namely the collection of logistic classification problems described by \cite{ChangL:2011} in the form of their library for support vector machines  (LIBSVM).
Finally we study a class of problems of much smaller scale, posing a different challenge in that for these problems it is inherently impossible to compute the cost function and the gradient exactly despite their small-scale nature. 
In our experiments we compare against relevant state-of-the-art methods.
All experiments were run on a MacBook Pro 2.8GHz laptop with 16GB of RAM using Matlab 2017b.
More details about some of the experiments and their background are available in the supplemental material.

\subsection{Synthetic example -- Rosenbrock's banana function}
\label{sec:SE}
\begin{figure}[hb!]
  \centering
  \begin{subfigure}[h!]{0.45\columnwidth}
      \includegraphics[width=\columnwidth]{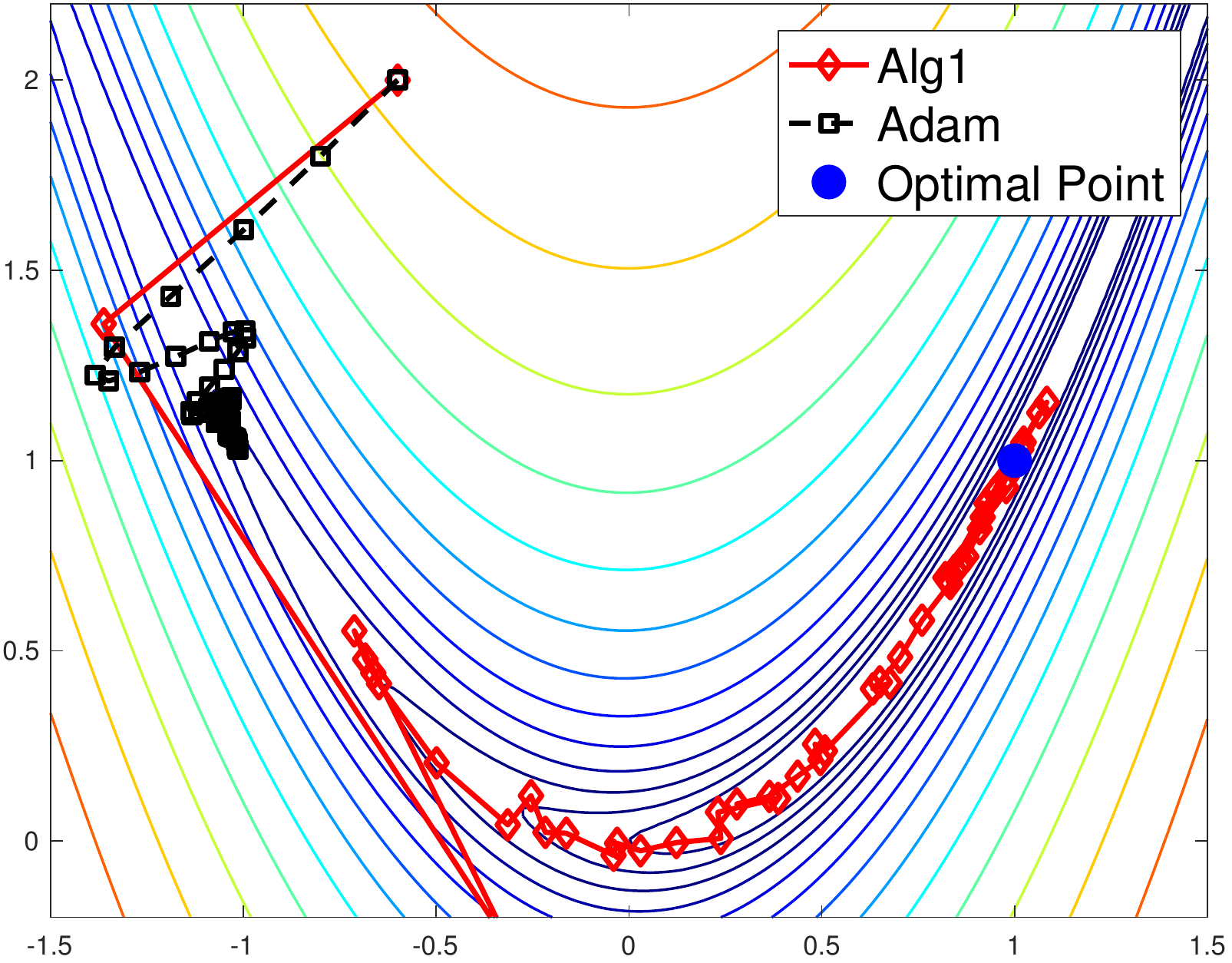}
    \caption{First $50$ iterates.}
    \label{fig:contour}
  \end{subfigure}
 \,\,\, \begin{subfigure}[h!]{0.45\columnwidth}
    \includegraphics[width=\columnwidth]{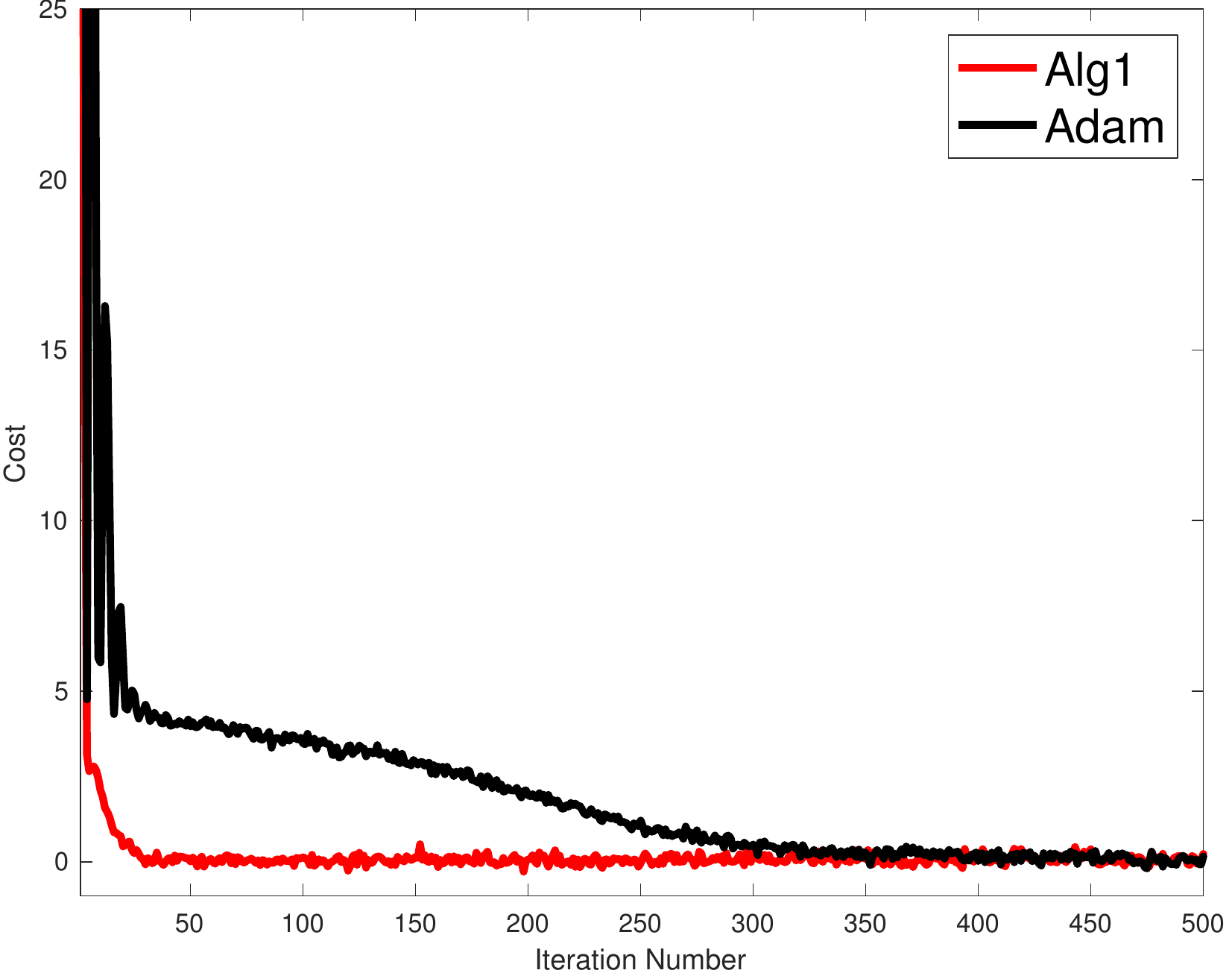}
    \caption{Cost per iteration}
    \label{fig:bananaCost}
  \end{subfigure}
  \caption{Rosenbrock's banana function. Figure (a) shows the
    contour lines of the cost function together with 50 iterates from 
    Algorithm~\ref{alg:SQN} and Adam, respectively. Figure (b) shows the cost 
    per iteration for the same two algorithms.}
\end{figure}

Let us start by demonstrating our proposed algorithm on a simple and possibly 
familiar problem, namely, that of minimising the Rosenbrock banana function 
(a contour plot of the Rosenbrock function is provided in Figure~\ref{fig:contour}). 
To emulate the stochastic nature of the problems considered in this paper, we 
have added artificial noise (standard deviation of $\sigma = 0.1$) to both the cost 
function and gradient calculations.

The Rosenbrock function is well-known to cause difficulty for
first-order methods because the Hessian matrix has disparate
eigenvalues along its banana-shaped valley. To compare our approach,
we also implemented the Adam algorithm from
\cite{KingmaB:2015}. Figure~\ref{fig:contour} shows the first~$50$
iterates of both methods. Clearly the proposed algorithm is converging
to a region around the optimal point while Adam is making slower
progress along the valley.  Figure~\ref{fig:bananaCost} shows the cost
value as a function of iteration, and while both methods converge to a
similar cost value, the proposed approach achieves this quite quickly.

While it is difficult and ill-advised to draw strong conclusions from this 
tiny experiment, it does provide some confidence that the second-order
information is indeed captured and exploited by our proposed algorithm.
% \ts{pedagogical illustration.}
% \ts{Show the improvement over ADAM, ADAgrad, and others in this example. That would be a nice indication.}
%\def\Put(#1,#2)#3{\leavevmode\makebox(0,0){\put(#1,#2){#3}}}
%=============================================================================
%===============================  New Section  ===============================
%=============================================================================
\subsection{MNIST}
\label{sec:MNIST}
Deep convolutional neural networks (CNNs) with multiple layers of
convolution, pooling and nonlinear activation functions are delivering
state-of-the-art results on many tasks in computer vision. We are here
borrowing the stochastic optimisation problems arising in using such a
deep CNN to solve the MNIST benchmark.  The particular CNN structure
used in this example employs $5\times 5$ convolution kernels,
pooling layers and a fully connected layer at the end. We made use of the publicly available code provided
by~\cite{Zhang:2016}, which contains all the implementation details.
In Figure~\ref{fig:MNIST} we show the average cost versus time for
$20$ Monte-Carlo trials with Algorithm~\ref{alg:SQN} (with $b=300$, $m=30$ and $\lambda = 0.1$),
Adam developed by \cite{KingmaB:2015} and the basic SG
algorithm. Note that the three algorithms all make use of the same
gradients.
%\aw{Worth noting, I tried the SGD step-length reduction instead of the and both work very well.}
%=============================================================================
%===============================  New Section  ===============================
%=============================================================================
%\subsection{Logistic classification problems}
\subsection{Logistic loss and a 2-norm regularizer}
\label{sec:LIBSVM}
%A collection of logistic classification problems from the LIBSVM library by \cite{ChangL:2011}. These problems are commonly used for profiling optimisation algorithms of the kind introduced in this paper, facilitating comparison with existing state of the art algorithms;
The task here is to solve seven different empirical risk minimisation problems using a logistic loss function with an L2
regularizer. The data is taken from \cite{ChangL:2011}. These
problems are commonly used for profiling optimisation algorithms of
the kind introduced in this paper, facilitating comparison with
existing state-of-the-art algorithms. More specifically, we have used
the same set-up as \cite{GowerGR:2016}, which inspired this study. A
summary of the salient features of each problem is provided in
Table~\ref{tab:table}. Recall that our algorithm only requires the
user to select two tuning parameters, namely the mini-batch size used
($b$), and the memory length ($m$). Our choices for these parameters
are listed in Table~\ref{tab:table}.

\begin{table}[h]\centering
	\small{
		\ra{1.2}        			% Controls the space between cells.
		\begin{tabular}{lll|lll}
			\toprule
			\textbf{Problem} & $n$ & $d$ & $b$ & $m$ & $\lambda$\\
			\midrule
			\texttt{gisette} & $6\thinspace000$ & $5\thinspace000$ & $500$ & $20$ & $1.0$\\
			\texttt{covtype} & $581\thinspace012$ & $54$ & $763$ & $54$ & $0.04$\\
			\texttt{HIGGS} & $11\thinspace000\thinspace000$ & $28$ & $3\thinspace317$ & $28$ & $0.04$\\
			\texttt{SUSY} & $3\thinspace548\thinspace466$ & $18$ & $5\thinspace000$ & $18$ & $0.04$\\
			\texttt{epsilon} & $400\thinspace000$ & $2\thinspace000$ & $1\thinspace000$ &$20$ & $0.2$\\
			\texttt{rcv1} & $20\thinspace242$ & $47\thinspace236$ & $284$ & $2$& $0.2$\\
			\texttt{URL} & $2\thinspace396\thinspace130$ & $3\thinspace231\thinspace961$ & $1\thinspace798$ & $50$& $0.04$\\
			\bottomrule
		\end{tabular}
		\caption{List of seven problems (columns 1), the number of data
			points~$n$ (column 2), the number of variables $d$ (column 3), the mini-batch size~$b$ (column 4), the memory size~$m$ (column 5), and the regulariser~$\lambda$ (column 6).}
		\label{tab:table}
	}
\end{table}

\begin{table}[h]
	\small{
		\centering
		\ra{1.2}        			% Controls the space between cells.
		\begin{tabular}{lllll}
			\toprule
			\textbf{Problem} & Alg1  & MNJ & GGR & SVRG \\
			\midrule
			\texttt{gisette} & \bf{0.005}  & 0.244 & 0.0176 & 0.172 \\
			\texttt{covtype} & \bf{0.514}  & 0.684 & \bf{0.514} & 0.667\\
			\texttt{HIGGS} & \bf{0.638}  & \bf{0.638} & \bf{0.638} & \bf{0.638}\\
			\texttt{SUSY} & \bf{0.458}  & \bf{0.458} & \bf{0.458} & \bf{0.458}\\
			\texttt{epsilon} & \bf{0.282}  & \bf{0.282} & \bf{0.282} & 0.421\\
			\texttt{rcv1} & \bf{0.202}  & \bf{0.202} & \bf{0.202} & 0.280 \\
			\texttt{URL} & 0.0196  & \bf{0.0193} & 0.0249 & 0.0639\\
			\bottomrule
		\end{tabular}
		\caption{Cost function values for each problem (columns 1),
			and each method Alg1 (column 2), MNJ (column 3), GGR (column 4) and SVRG (column 5). Minimum value in bold face.}
		\label{tab:table2}
	}
\end{table}

\begin{figure}[ht!]
	\centering
	\begin{subfigure}[h!]{0.35\columnwidth}
		\includegraphics[width=\columnwidth]{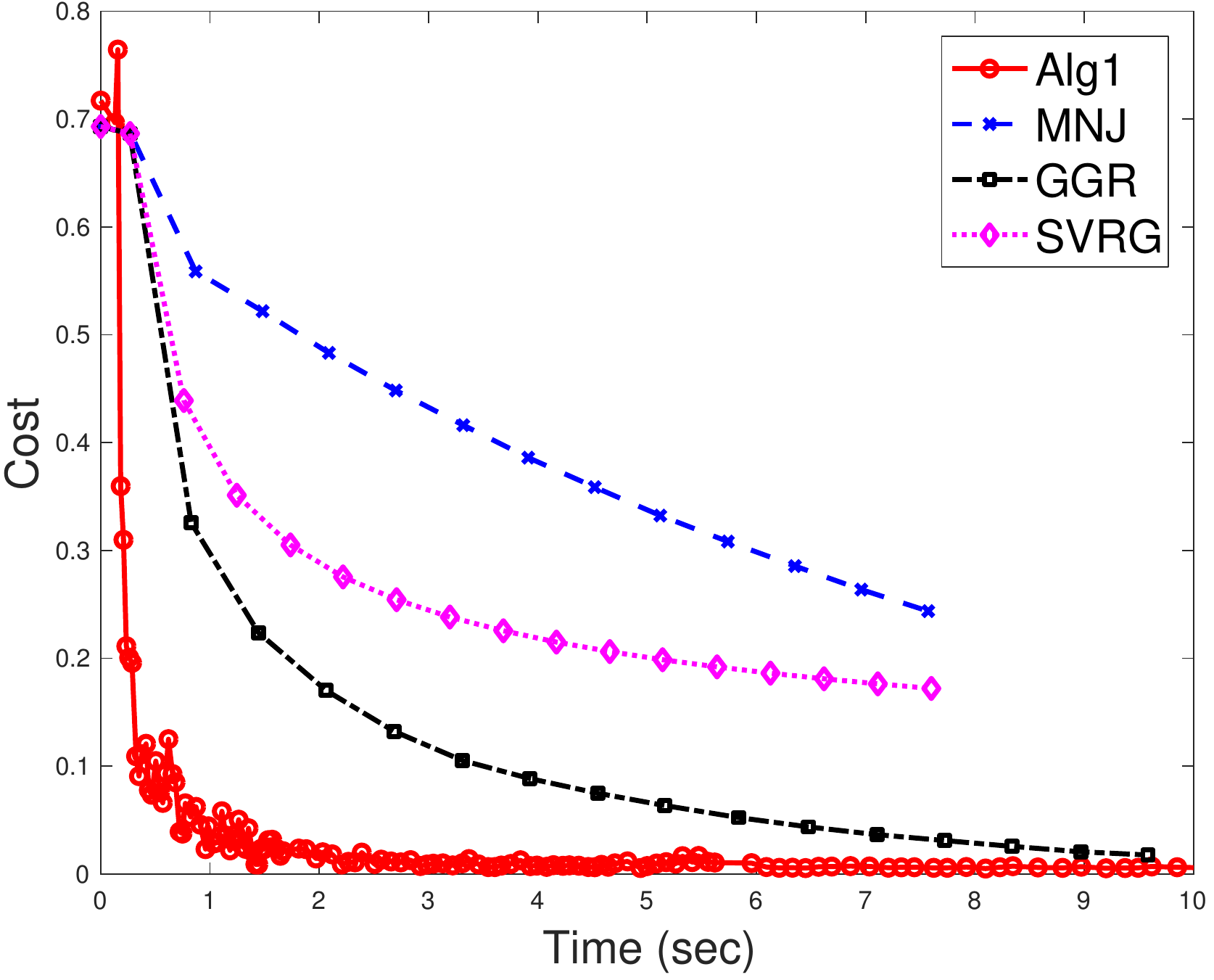}
		\caption{\texttt{gisette}}
		\label{fig:gisette}
	\end{subfigure}
	\begin{subfigure}[h!]{0.35\columnwidth}
		\includegraphics[width=\columnwidth]{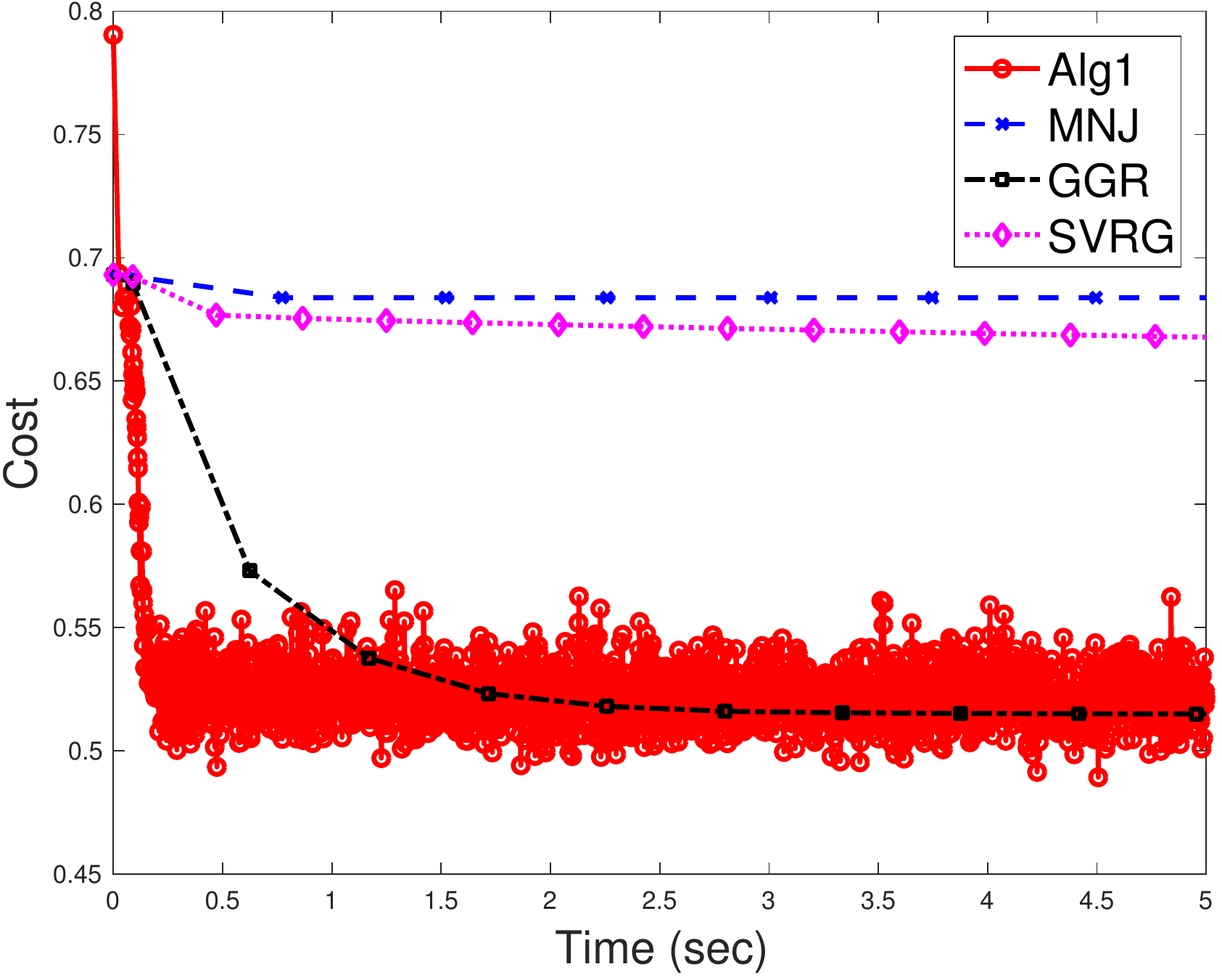}
		\caption{\texttt{covtype}}
		\label{fig:covtype}
	\end{subfigure}
	\begin{subfigure}[h!]{0.35\columnwidth}
		\includegraphics[width=\columnwidth]{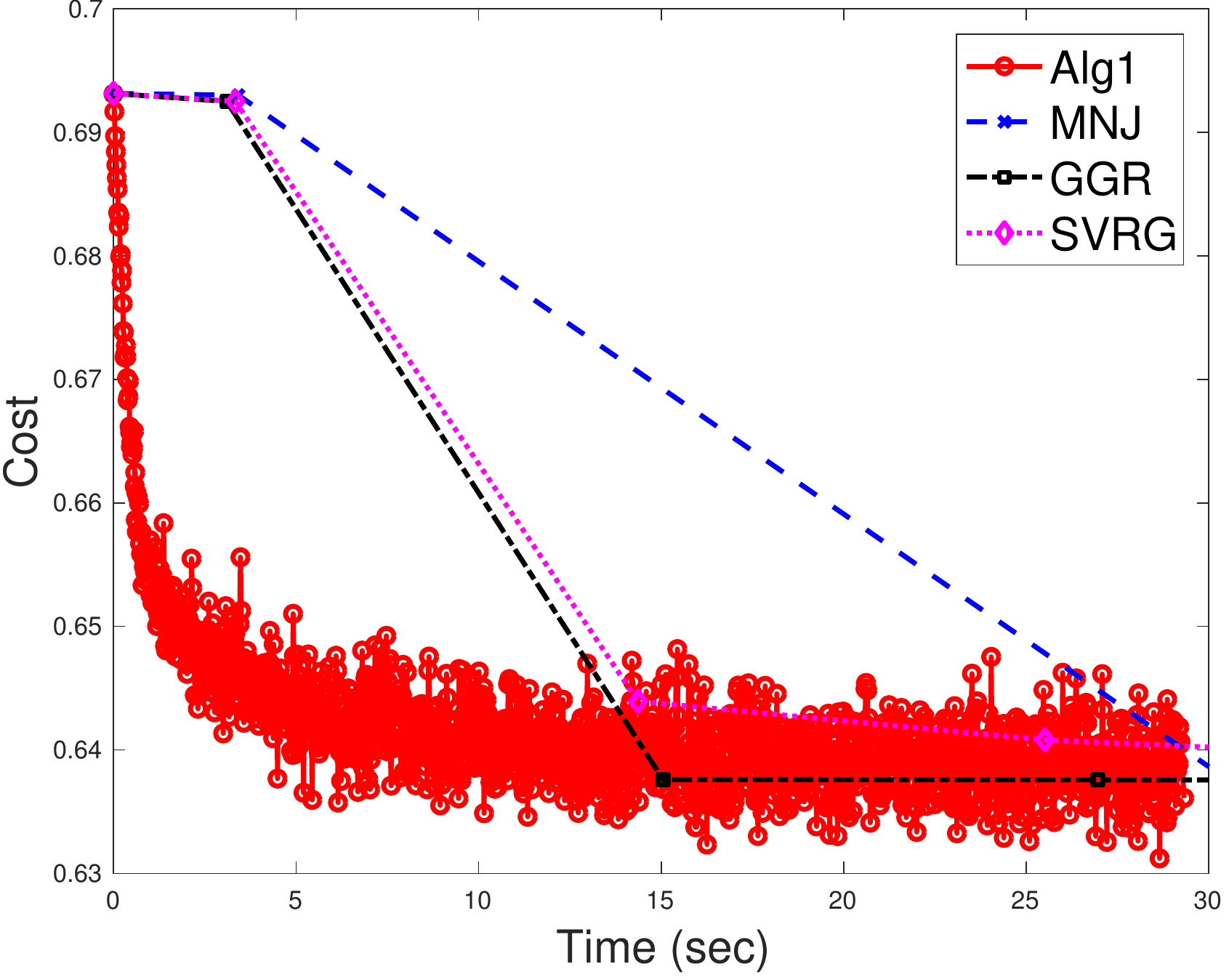}
		\caption{\texttt{HIGGS}}
		\label{fig:higgs}
	\end{subfigure}
	%   \caption{Logistic Classification Problems}
	% \end{figure} 
	% \begin{figure}[h!]
	%   \centering
	\begin{subfigure}[h!]{0.35\columnwidth}
		\includegraphics[width=\columnwidth]{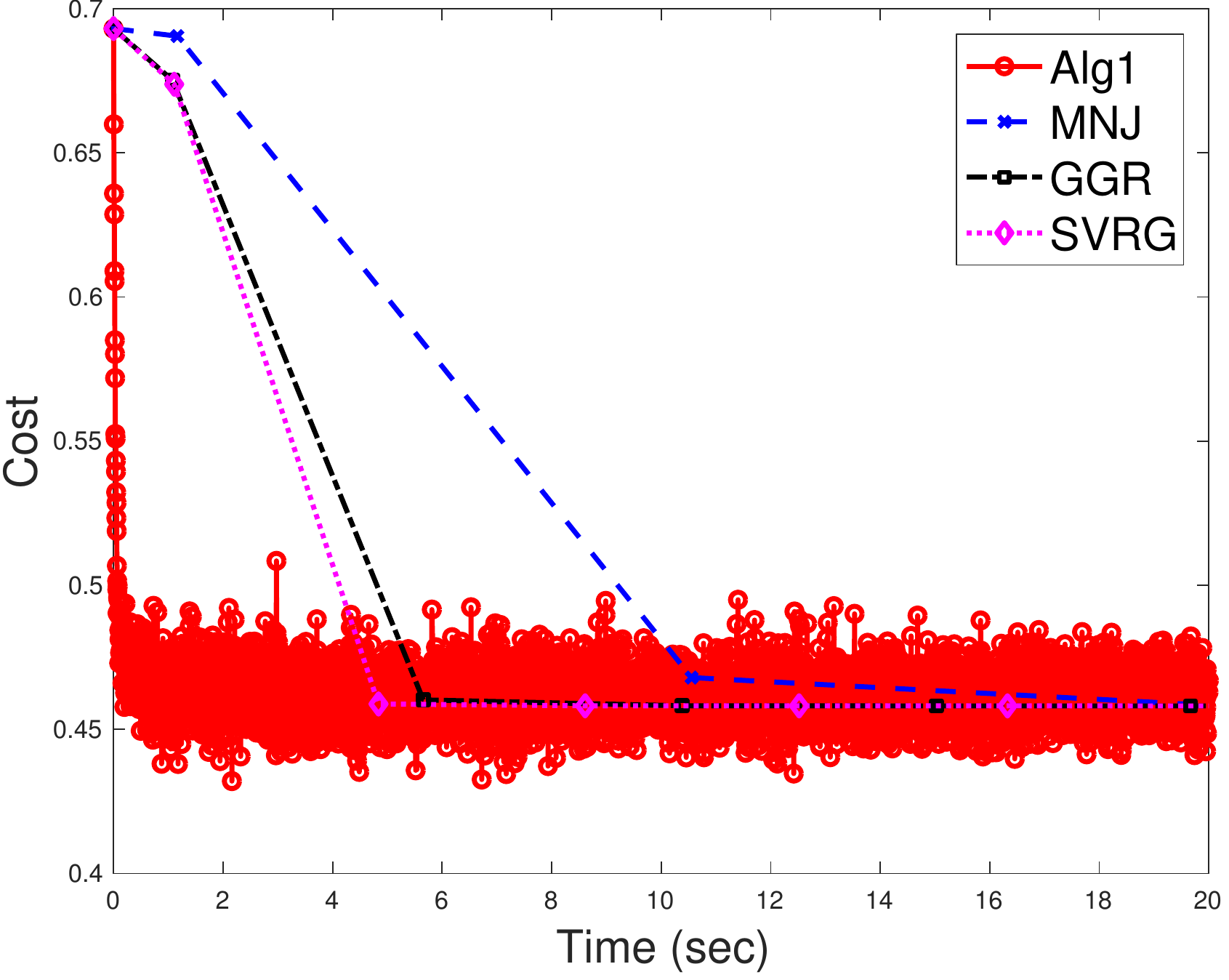}
		\caption{\texttt{SUSY}}
		\label{fig:susy}
	\end{subfigure}
	\begin{subfigure}[h!]{0.35\columnwidth}
		\includegraphics[width=\columnwidth]{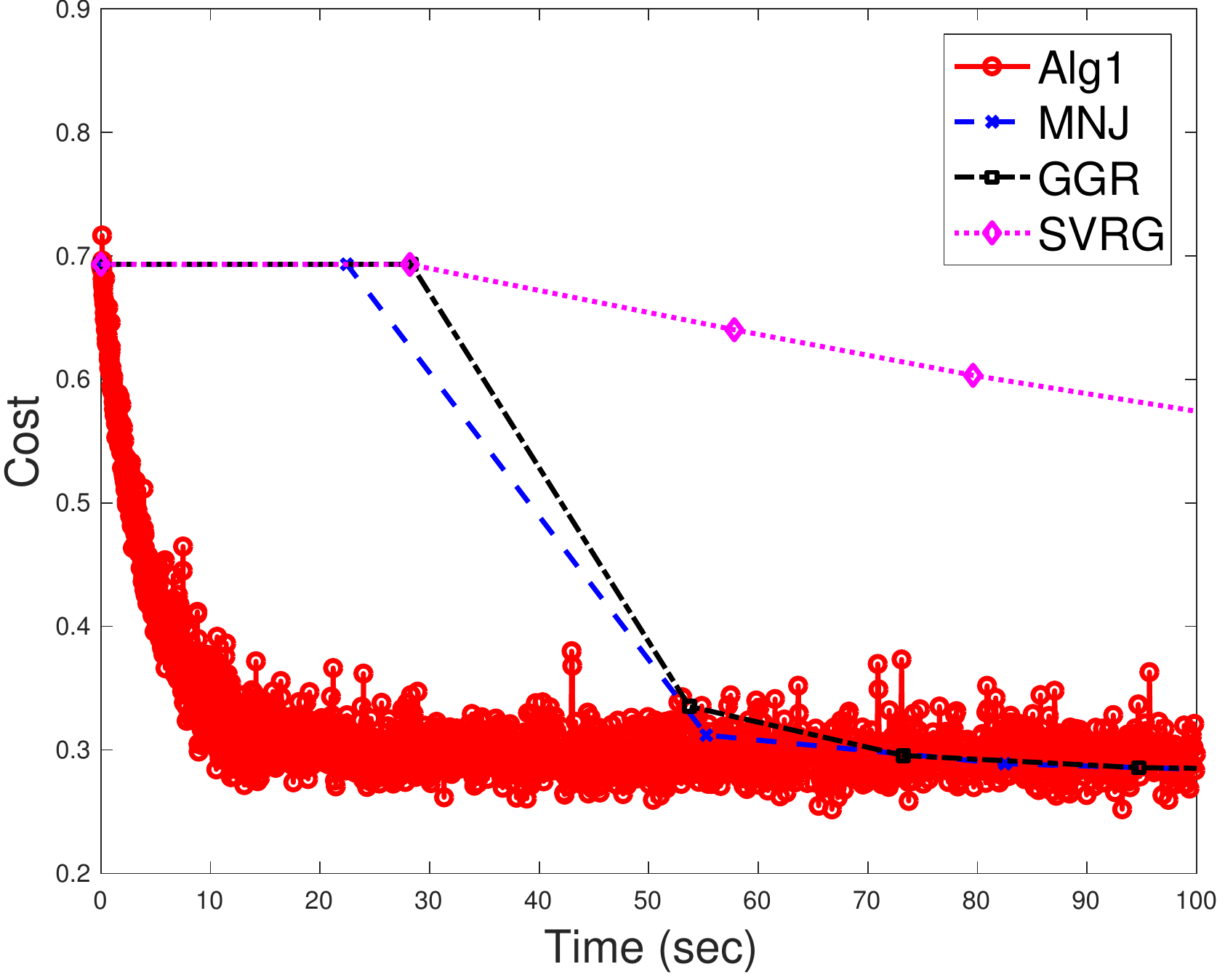}
		\caption{\texttt{epsilon}}
		\label{fig:epsilon}
	\end{subfigure}
	\begin{subfigure}[h!]{0.35\columnwidth}
		\includegraphics[width=\columnwidth]{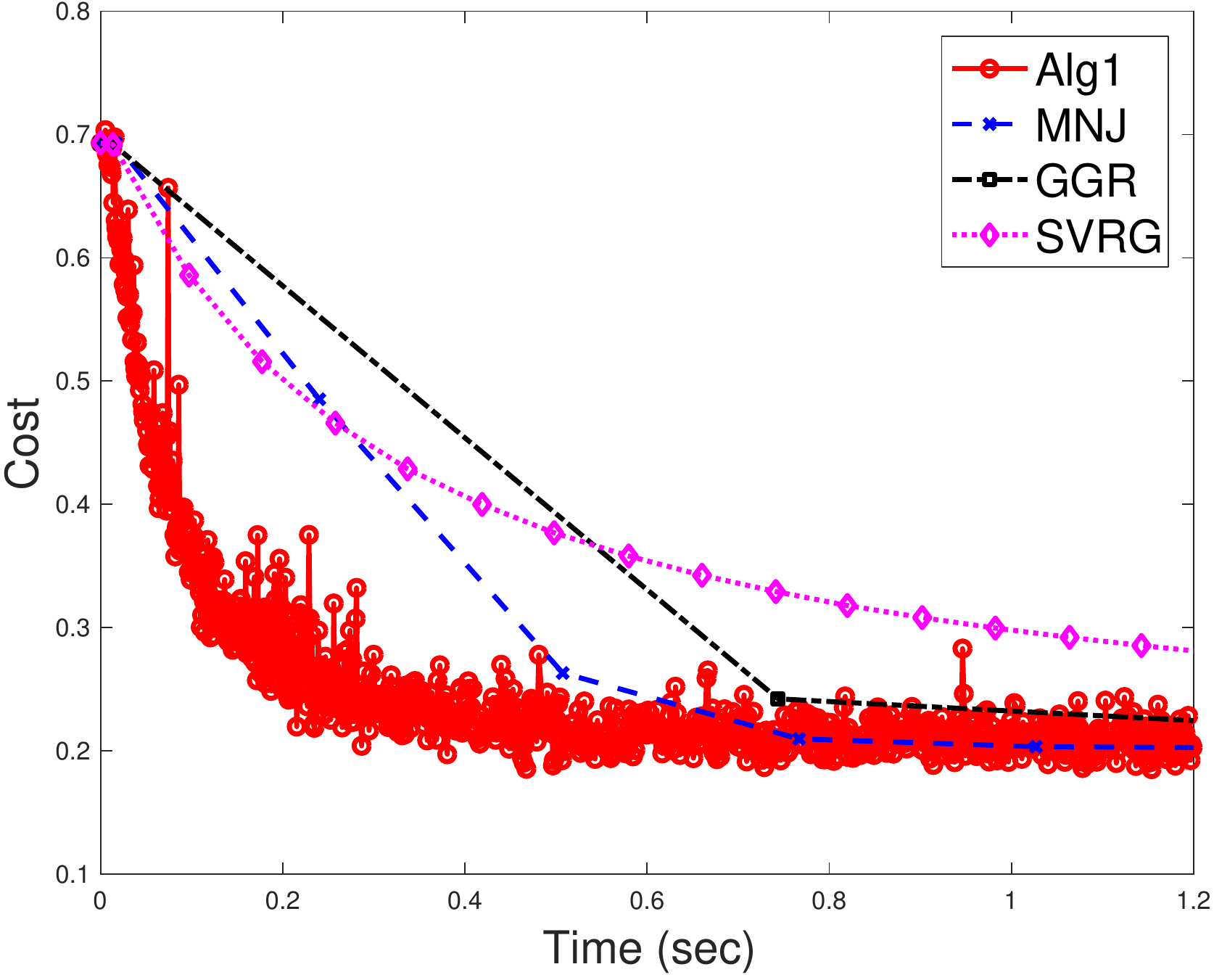}
		\caption{\texttt{RCV1}}
		\label{fig:rcv1}
	\end{subfigure}
	\begin{subfigure}[h!]{0.35\columnwidth}
		\includegraphics[width=\columnwidth]{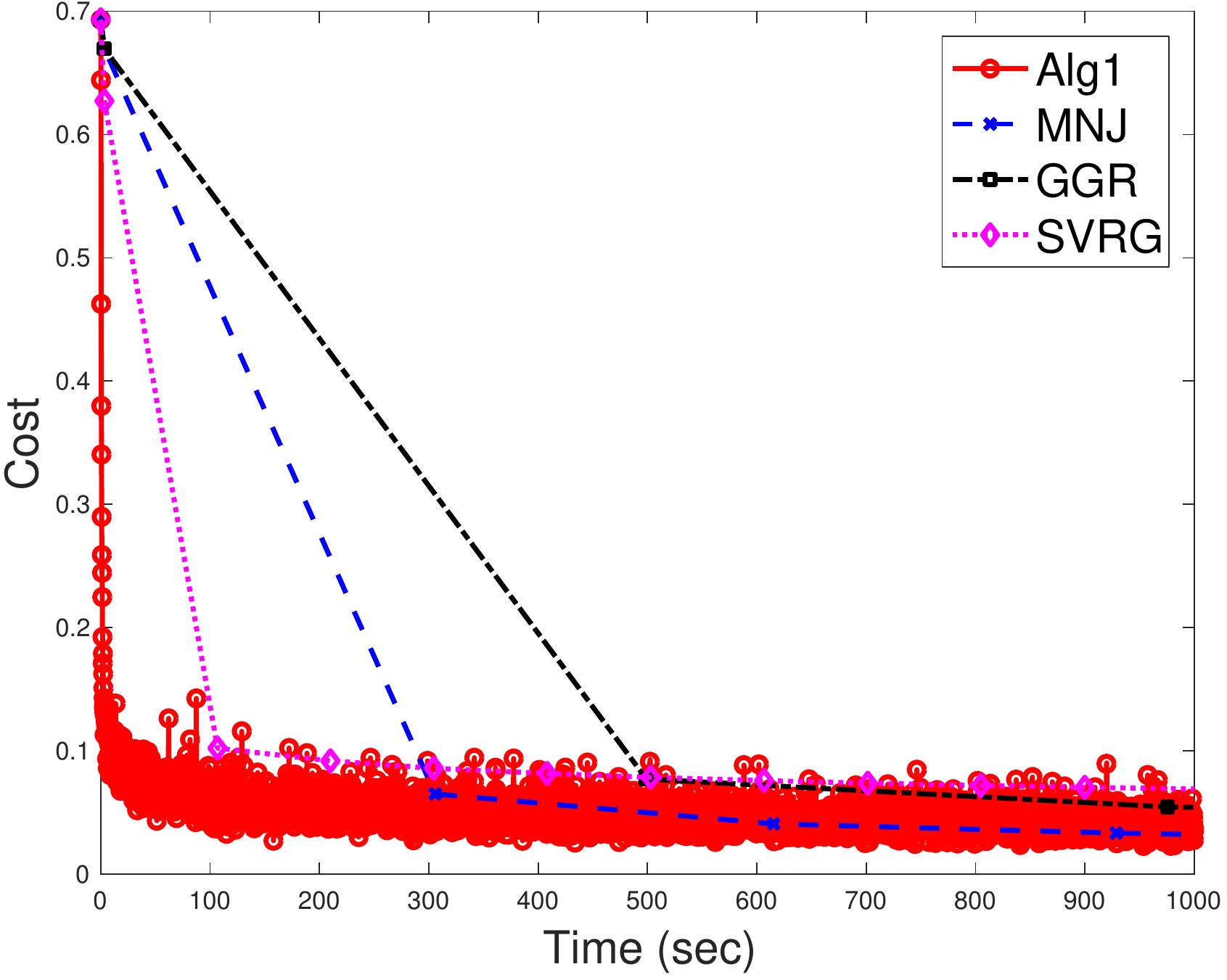}
		\caption{\texttt{URL}}
		\label{fig:url}
	\end{subfigure}
	\begin{subfigure}[h!]{0.35\columnwidth}
		%    \includegraphics[width=\columnwidth]{mnist_plot}
		%    \caption{\texttt{MNIST}}
		\includegraphics[width=\columnwidth]{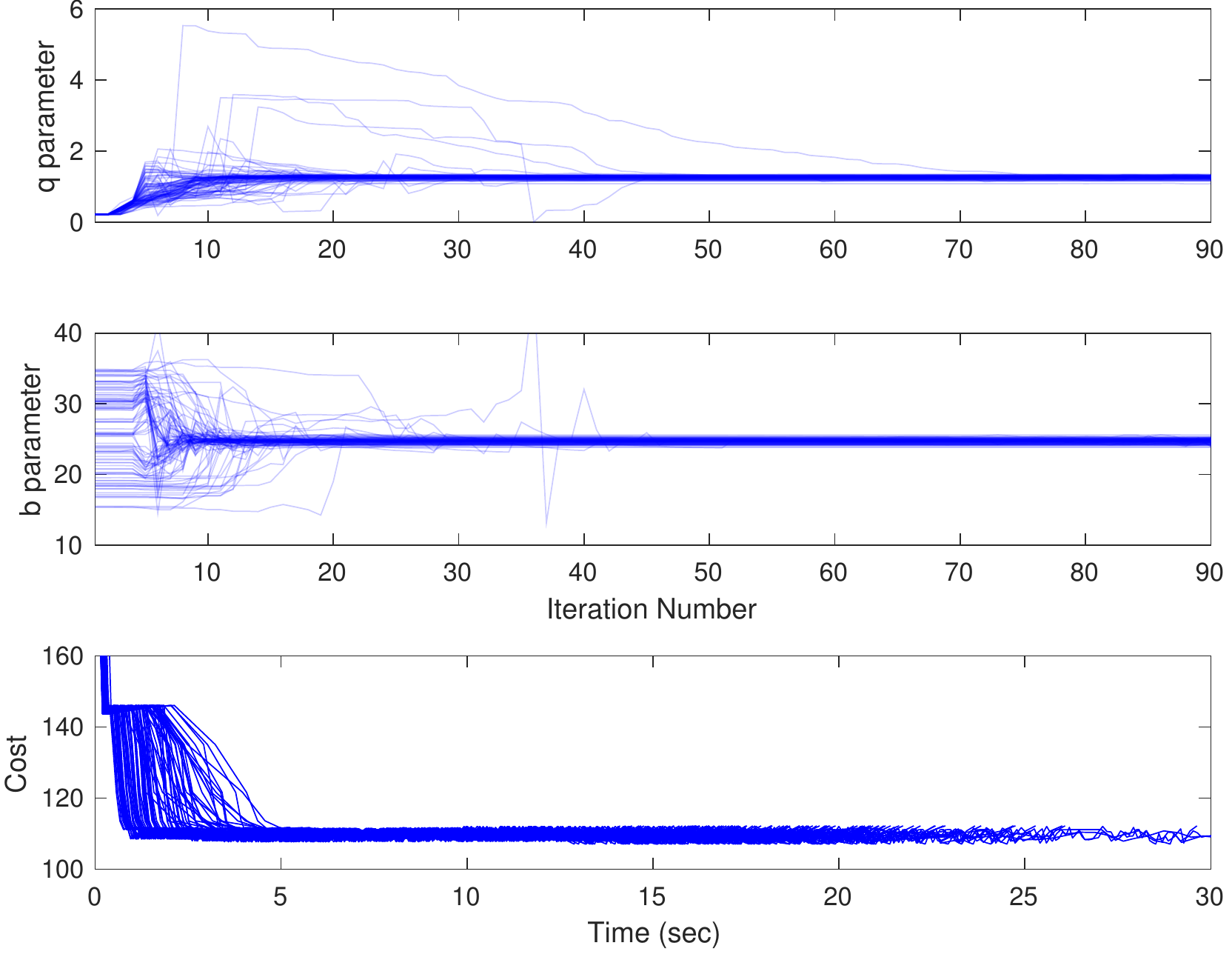}
		\caption{Nonlinear SSM.}
		\label{fig:NLSSM}
	\end{subfigure}
	\caption{Performance on seven classification tasks using a  logistic loss with a two-norm regulariser (Figures (a)--(g)). In Figure~(h) we show the result on a learning parameters in a challenging nonlinear dynamical system.}
	\label{fig:Res}
\end{figure}

We compared Algorithm~\ref{alg:SQN} (denoted as \texttt{Alg1})
against three existing
methods from the literature, namely, the limited memory stochastic
block BFGS method from \cite{GowerGR:2016} (denoted as \texttt{GGR})
and the limited memory stochastic BFGS method of \cite{MoritzNJ:2016}
(denoted as \texttt{MNJ}) and the stochastic variance reduced gradient
(SVRG) by \cite{JohnsonZ:2013} (denoted \texttt{SVRG}). For the
\texttt{GGR}, \texttt{MNJ} and \texttt{SVRG} approaches we used the
recommended tuning of each algorithm. In the case of \texttt{GGR} we
used the \texttt{prev} variant as this performed best across all test
problems\footnote{The implementation for \texttt{GGR} and \texttt{MNJ}
  was downloaded from
  \url{www.maths.ed.ac.uk/~prichtar/i_software.html}}. The result is illustrated in Table~\ref{tab:table2} and Figure~\ref{fig:Res}.

%=============================================================================
%===============================  New Section  ===============================
%=============================================================================
\subsection{Nonlinear system identification}\label{sec:SMC}
Another important application requiring stochastic optimisation
problems to be solved is that of nonlinear system identification,
where the task is to learn unknown parameters in nonlinear dynamical
systems (see Appendix~\ref{sec:app} for further details).  Here the
stochasticity arises due to the fact that it is impossible to exactly
evaluate the cost function (provided by maximum likelihood) and its
gradients.  Instead we have to resort to approximations resulting in
noisy evaluations of the kind~\eqref{eq:NoisyCostGrad}.  Consider the
problem of learning the parameters~$b$ and~$q$ for the following
nonlinear and time-varying state-space model,
\begin{subequations}  
	\label{eq:nlp}
	\begin{align}
	x_{t+1} &= 0.5x_t + b\frac{x_t}{1 + x_t^2} + 8\cos (1.2t) + q^{-1} w_t, \label{eq:54a}\\
	y_t &= 0.05x_t^2 + e_t,\label{eq:49}
	\end{align}
\end{subequations}
where the true parameters are $b^\star = 25$ and $q^\star = 1/\sqrt{0.5}$. The noise terms are mutually independent and given by $w_t\sim\N(0, 1)$ and $e_t \sim \N(0, 0.1)$.
%  \begin{align}
%  \begin{bmatrix}
%    w_t \\ 
%    e_t
%  \end{bmatrix} &\sim \mathcal{N} \left (\begin{bmatrix}
%      0 \\ 
%      0
%    \end{bmatrix}, \begin{bmatrix}
%      1 & 0 \\ 
%      0 & 0.1
%    \end{bmatrix}
%  \right)
%  \label{eq:55a}
%\end{align}
%\end{subequations}
This has been acknowledged as a challenging problem \citep{DoucetGA:2000,GodsillDW:2004} within the sequential Monte Carlo (SMC) community. 
%This example has previously been investigated by the current authors~\cite{SchonWN:2011} and is profiled again here due to it being acknowledged as a challenging problem~\cite{DoucetGA:2000,GodsillDW:2004}.
The results using 100 measurements and 200 particles for 100 Monte--Carlo simulations are provided in  Figure~\ref{fig:NLSSM}.

%\ts{Update Figure 2 with correct $q$? or update $q^{\star}$ above.}

%\begin{figure}[h!]
%  \centering
%  \includegraphics[width=0.95\columnwidth]{nlss_plot}
%  \caption{Result of using Algorithm~\ref{alg:SQN} in solving the stochastic optimisation problem allowing us to learn the unknown parameters~$b$ and~$q$ in~\eqref{eq:nlp}.}
%  \label{fig:nlss}
%\end{figure}

%=============================================================================
%===============================  New Section  ===============================
%=============================================================================
\section{Conclusion and future work}
\label{sec:Conc}
% Conclusions and future work
% 
In this paper we have developed a new approach for solving large-scale
stochastic optimisation problems by combining curvature information in
computing the search direction with the use of an adaptive step length
that is regulated by the cost function. The local curvature
information is captured using a limited memory method whose
computational cost scales linearly in the data size.
We demonstrate our approach on a range of problems from different
fields of research including a suite of challenging large-scale
problems. The proposed method performs well against state-of-the-art
techniques and we believe that this provides some impetus for further
research.
%warrants further investigation. %Another area to investigate concerns the use of variance reduction techniques, which has been ignored in the current work.
As a final remark, an interesting situation occurs when we employ
Algorithm~\ref{alg:SQN} with $\rho = 0$ together with a decaying
maximum step length~$\bar{\alpha}_k$. In the limit, this mimics SG
methods, but in early iterations it regulates the step length in order
to reduce the cost. This circumvents the requirement of conservative
initial step lengths.

%=============================================================================
%===============================  New Section  ===============================
\section*{Acknowledgements}
We would like to thank the participants of the Sydney control
conference 2017 for very useful discussion and feedback on a
presentation leading up to this work. We would also like to thank
Fredrik Lindsten, Johan Dahlin and Jack Umenberger for very useful
comments on an early draft of this paper.  This research was
financially supported by the Swedish Foundation for Strategic Research
(SSF) via the project \emph{ASSEMBLE} (contract number: RIT15-0012)
and the Swedish Research Council via the projects \emph{Learning
  flexible models for nonlinear dynamics} (contract number:
2017-03807) and \emph{NewLEADS - New Directions in Learning Dynamical
  Systems} (contract number: 621-2016-06079).

%=============================================================================
%===============================  New Section  ===============================
%=============================================================================
\section{Appendix -- Learning nonlinear dynamical systems}
\label{sec:app}
%%=============================================================================
%%===============================  New Section  ===============================
%%=============================================================================
%\section{}
%\label{sec:SMnls}
%=============================================================================
%===============================  New Section  ===============================
%=============================================================================
\subsection{Problem formulation}
%To make this very concrete, let us study the problem of maximum likelihood learning of nonlinear state-space models
Consider the following general nonlinear state-space model
\begin{subequations}
	\begin{align}
	x_{t} &= f(x_{t-1},\theta) + w_t,\\
	y_t &= h(x_t, \theta) + e_t,
	\end{align}
\end{subequations}
where $x_t$ denotes the state, $y_t$ denotes the measurement
and~$\theta$ denotes the unknown (static) parameters. The two
nonlinear functions $f(\cdot)$ and $h(\cdot)$ denotes the nonlinear
functions describing the dynamics and the measurements,
respectively. The process noise is Gaussian distributed with zero mean
and covariance $Q$, $w_t\sim \N(0, Q)$ and the measurement noise is
given by $e_t\sim\N(0, R)$. Finally, the initial state is distributed
according to $x_0 \sim p(x_0\mid\theta)$. The problem we are
interested in is to estimate the unknown parameters~$\theta$ by making
use of the available measurements
$y_{1:\T} = \{y_1, y_2, \dots, y_{\T}\}$ to maximize the likelihood
function $p(y_{1:\T} \mid \theta)$
\begin{align}
\label{eq:MaxLik}
\max_{\theta}{p(y_{1:\T} \mid \theta)}.
\end{align}
In the supplemental material we provide more background on how to
compute approximations of the likelihood function~\eqref{eq:MaxLik}
and its gradients using sequential Monte Carlo (SMC) methods
\citep{Gordon:1993,Kitagawa:1993}. For a tutorial introduction to SMC
methods we refer to \cite{DoucetJ:2011} and their use in solving
system identification problems is offered by \cite{SchonLDWNSD:2015}
and \cite{Kantas:2015}.

%=============================================================================
%===============================  New Section  ===============================
%=============================================================================
\subsection{Computing the likelihood and its gradient}
The likelihood function can via repeated use of conditional probabilities be rewritten as 
\begin{align}
p(y_{1:\T}\mid \theta) = \prod_{t=1}^{\T} p(y_t\mid y_{1:t-1},\theta),
\end{align}
with the convention that $y_{1:0} = \emptyset$. The one step ahead predictors are available via marginalization
\begin{align}
p(y_t\mid y_{1:t-1}, \theta) = \int p(y_t, x_t \mid y_{1:t-1}, \theta) \myd x_t 
= \int p(y_t\mid x_t, \theta)  p(x_t\mid y_{1:t-1}, \theta) \myd x_t.
\end{align}
One intuitive interpretation of the above integral is that it corresponds to averaging over all possible values for the state $x_{t}$. The challenge is of course how to actually compute this integral. By making use of particle filter \citep{Gordon:1993,Kitagawa:1993} to approximate the likelihood we are guaranteed to obtain an unbiased estimate~\citep{DelMoral:2004}.

The likelihood gradients can also be computed using particle filters, for example by making use of \emph{Fisher's identity} \citep{CappeMR:2005}
\begin{align}
\nabla_{\theta}\ell(\theta)\big|_{\theta = \theta_k} = 
\nabla_{\theta}\mathcal{Q}(\theta,\theta_k)\big|_{\theta = \theta_k} 
\end{align}
where we have defined
\begin{subequations}
	\begin{align}
	\ell(\theta) &= \ln p(y_{1:\T}\mid \theta),\\
	\mathcal{Q}(\theta,\theta_k) &= 
	\int \ln p(x_{0:\T}, y_{1:\T} \mid \theta) p(x_{0:\T}\mid y_{1:\T},  \theta_k) \myd x_{0:\T}.
	\end{align}
\end{subequations}
The particle filter---which is one member of the family of sequential Monte Carlo (SMC) methods---has a fairly rich history when it comes to solving nonlinear system identification problems. For an introductory overview we refer to~\cite{SchonLDWNSD:2015,Kantas:2015}.

The likelihood and its gradient cannot be calculated exactly in this
case and we therefore employed sequential Monte Carlo methods and
Fisher's identity \citep{CappeMR:2005,NinnessWS:2010} to provide noisy
estimates of both.  The number of particles used to calculate these
terms was 500 in all cases.  Note that each simulation required no
more than 8 seconds of computation time on a MacBook Pro 2.8GHz Intel
i7.

\bibliographystyle{apalike}
\bibliography{bibmain}

\end{document}